\documentclass[journal]{IEEEtran}
\usepackage{graphics} % for pdf, bitmapped graphics files
\usepackage{amsmath} % assumes amsmath package installed
\usepackage{cite}
\usepackage{subfigure}
\usepackage{tabulary}
\usepackage{multirow}
\usepackage{todonotes}
\graphicspath{{figs/}}
\usepackage[colorlinks,linkcolor=blue,anchorcolor=blue,citecolor=red]{hyperref} 
\usepackage{wrapfig,lipsum,booktabs}
\usepackage{color}
\usepackage{cite}
\usepackage{algorithmic} %format of the algorithm 
\usepackage[ruled,vlined,commentsnumbered]{algorithm2e}

\usepackage[flushleft]{threeparttable}
\usepackage{gensymb}  % degree symbol
\usepackage{amsmath}
\usepackage{amssymb}

\usepackage{flushend}
\usepackage{balance}

\graphicspath{{figs/}}

\usepackage{siunitx}
\usepackage{url}   % url link

\begin{document}

\markboth{IEEE Robotics and Automation Letters. Preprint Version. Accepted June, 2019}
{Yang \MakeLowercase{\textit{et al.}}: ShortTitle}

\title{\LARGE \bf
Monocular Object and Plane SLAM in Structured Environments
}

\author{Shichao Yang, Sebastian Scherer % <-this % stops a space
\thanks{Manuscript received February 24, 2019; revised May 6, 2019; accepted June 7, 2019. This paper was recommended for publication by Editor Cyrill Stachniss upon evaluation of the reviewers' comments. The work was supported by the Amazon Research Award \#2D-01038138. (\textit{Corresponding author: Shichao Yang})}
\thanks{The authors are with the Robotics Institute, Carnegie Mellon University, Pittsburgh, PA, USA. Email of first author: \{shichaoy@andrew.cmu.edu, 2013ysc@gmail.com\}; Second author: basti@andrew.cmu.edu}
\thanks{This paper has supplementary downloadable multimedia material available at http://ieeexplore.ieee.org. The enclosed video demonstrates SLAM experimental results.}
\thanks{Digital Object Identifier (DOI): see the top of this page.}

%\thanks{The authors are with the Robotics Institute, Carnegie Mellon University, Pittsburgh, PA, USA. Email of first author: \{shichaoy@andrew.cmu.edu, 2013ysc@gmail.com\}; Second author: basti@andrew.cmu.edu}
%\thanks{The work was supported by the Amazon Research Award \#2D-01038138.}
%\thanks{Digital Object Identifier 10.1109/TRO.2019.2909168}

}

\maketitle

\begin{abstract}
    % importance of research, lack of existing approach
% Problem descripttion, scope of research
% Approach
% Results
In this paper, we present a monocular Simultaneous Localization and Mapping (SLAM) algorithm using high-level object and plane landmarks. The built map is denser, more compact and semantic meaningful compared to feature point based SLAM. We first propose a high order graphical model to jointly infer the 3D object and layout planes from single images considering occlusions and semantic constraints. The extracted objects and planes are further optimized with camera poses in a unified SLAM framework. Objects and planes can provide more semantic constraints such as Manhattan plane and object supporting relationships compared to points. Experiments on various public and collected datasets including ICL NUIM and TUM Mono show that our algorithm can improve camera localization accuracy compared to state-of-the-art SLAM especially when there is no loop closure, and also generate dense maps robustly in many structured environments.
\end{abstract}


\IEEEpeerreviewmaketitle

%\thispagestyle{empty}
%%%%%%%%%%%%%%%%%%%%%%%%%%%%%%%%%%%%%%%%%%%%%%%%%%%%%%%%%%%%%%%%%%%%%%%%%%%%%%%%

\section{Introduction}
\label{sec:object intro}
% what's the problem, importance of approach.
% briefly mention exisitng approches.
% proposed method, contributions.
\IEEEPARstart{S}{emantic} understanding and SLAM are two fundamental problems in computer vision and robotics. In recent years, there has been great progress in each field. For example, with the popularity of Convolutional Neural Network (CNN), the performance of object detection \cite{redmon2016yolo9000}, semantic segmentation \cite{badrinarayanan2017segnet}, and 3D understanding\cite{lee2017roomnet} has been improved greatly. In SLAM or Structure from Motion (SfM), approaches such as ORB SLAM \cite{mur2015orb} and DSO \cite{engel2017direct} are widely used in autonomous robots and Augmented Reality (AR) applications. However, the connections between visual understanding and SLAM are not well explored. Most existing SLAM methods represent the environments as sparse or semi-dense point cloud, which may not satisfy many applications. For example in autonomous driving, vehicles need to be detected in 3D space for safety and in AR applications, 3D objects and layout planes also need to be localized for virtual interactions.
%Objects and planes are important components especially in man-made environments. 

\begin{figure}[t]
  \centering
   \includegraphics[scale=0.16]{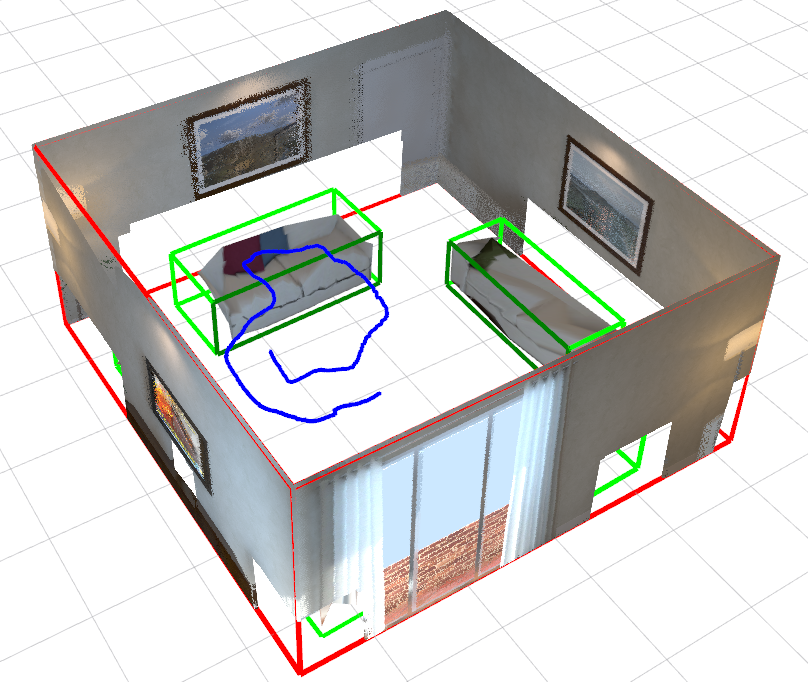}
   \includegraphics[scale=0.20]{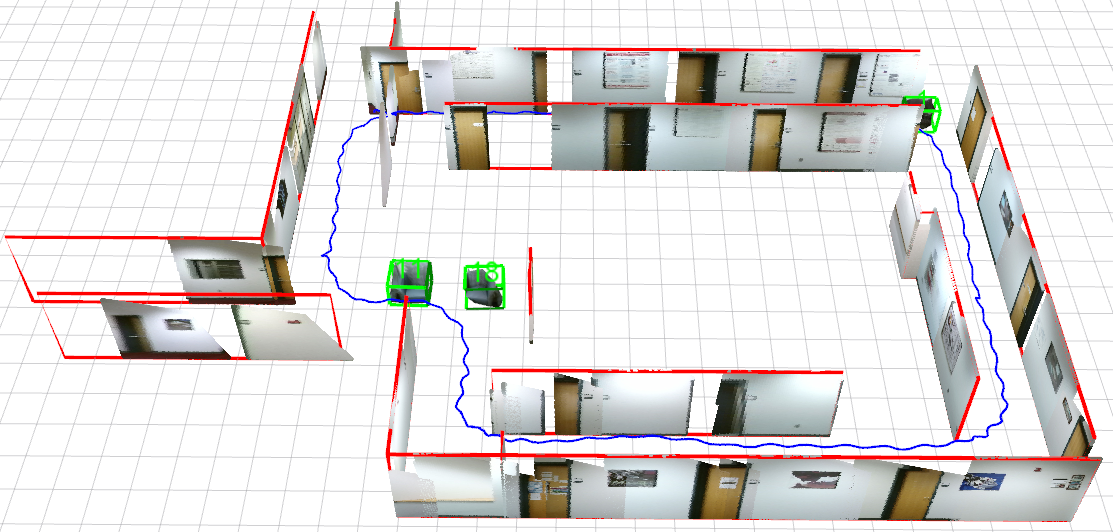}   
   \caption{Example result of dense SLAM map with points, objects (green box), planes (red rectangle) reconstructed using only a monocular camera. (top) ICL living room dataset.
   (bottom) Collected long corridor dataset.
   }
   \label{fig:intro figure}
\end{figure}
%[width=2.8in,height=1.6in]

There are typically two categories of approaches to combine visual understanding and SLAM. The decoupled approach first builds the SLAM point cloud then further labels \cite{kundu2014joint} \cite{yang2017semantic} or detects 3D objects \cite{pillai2015monocular} and planes \cite{bao2014understanding}, while the coupled approach jointly optimizes the camera pose with object and plane location. Most existing object SLAM \cite{salas2013slam++} \cite{galvez2016real} requires prior object models to detect and model objects, which limits the application in general environments.  Some prior works also utilize architectural planes for dense 3D reconstruction but mostly rely on RGBD \cite{lee2017joint} or LiDAR scanner \cite{xiao2014reconstructing}.

%In this paper, we develop a coupled method to demonstrate that high level object and plane landmarks can improve both camera pose estimation and dense mapping.

In this work, we propose a monocular object and plane level SLAM, without prior object and room shape models. It is divided into two steps. The first step is single image 3D structure understanding. Layout plane and cuboid object proposals are generated and optimized based on geometric and semantic image features. The second step is multi-view SLAM optimization. Planes and objects are further optimized with camera poses and point features in a unified bundle adjustment (BA) framework. Objects and planes provide additional semantic and geometric constraints to improve camera pose estimation as well as the final consistent and dense 3D map. Accurate SLAM pose estimation on the other hand improves the single image 3D detection. In summary, our contributions are as follows:

\begin{itemize}
\item A high order graphical model with efficient inference for single image 3D structure understanding.
\item The first monocular object and plane SLAM, and show improvements on both localization and mapping over state-of-the-art algorithms.
\end{itemize}

In the following, we first introduce the related work and single image 3D understanding in Sec \ref{sec:single image}, then explain multi-view SLAM optimization in Sec \ref{sec:slam}, followed by experiments in Sec \ref{sec:experiment}.

\section{Related Work}
\label{sec:object review}
%The related work is addressed in the two components of our work: 3D object detection and object SLAM.

%We propose to jointly optimize points, objects, and camera poses in unified bundle adjustment (BA) framework while existing approaches usually decouple them and might fail in some challenging scenarios. 
%Compared with the exhaustive sampling of 3D hypothesis \cite{chen2016monocular}, our method can greatly reduce 3D proposal numbers.

%Many existing usually first build SLAM/sfm point cloud then post processes to detect or select 3D objects, which might fail if the prior step SLAM fails in some challenging scenarios.
%Since we don't have prior object mode\textbf{•}ls, we need to parameterize object six DoF pose and dimensions.

%Objects can also provide depth and scale initialization for SLAM system. Without the prior model, we extend the commonly used 6 Dof object parameterization to 9 dimension with additional 3 dof for object sizes.

\subsection{Single image understanding}

The classic 3D object detection depends on hand-crafted features such as edge and texture \cite{lim2013parsing}. CNNs are also used to directly predict object poses from images \cite{tekin2017real}. For layout detection, the popular room model based on vanishing point is proposed by Hedau \textit{et al}\cite{hedau2009recovering}. There are also some CNN learning based approaches including \cite{ren2016coarse} and RoomNet \cite{lee2017roomnet}. Most of them only apply to the restricted four-wall Manhattan room models and are not suitable for general indoor environments.

%Therefore it cannot be directly used as SLAM landmark which should be invariant across frames such as corner points and line segments.

%but they are not suitable for SLAM landmark optimization because CNN prediction may be inconsistent across frames shown in Fig. \ref{fig:other methods not good}. 

For the joint 3D understanding of object and planes, Most works mostly utilize RGBD camera and cannot run in real time \cite{lin2013holistic}. More recent works directly predict the 3D position of objects and planes utilizing deep networks \cite{huang2018cooperative}.

%Their positions are optimized based on the spatial and semantic relationships such as occlusion, intersection, and concurrence

\iffalse
\begin{figure}[t]
\centering
\subfigure[]{
 \includegraphics[width=1.08in,height=0.9in]{0720_layout_vis.jpg}
 \label{fig:plane proposal sample} 
}\hspace{-0.8em}
\subfigure[]{
 \includegraphics[width=1.08in,height=0.9in]{0740_layout_vis.jpg}
 \label{fig:object proposal sample}
}\hspace{-0.8em}
\subfigure[]{
 \includegraphics[width=1.08in,height=0.9in]{cfile_layout.png}
 \label{fig:object proposal sample}
}
\caption{Related work of deep learning based layout detection \cite{ren2016coarse}. In (a-b), they directly predict the layout edge probability shown as red color. For similar input images (a-b), they generate different edge masks. In (c) red line is predicted layout and green is ground truth. We can see that some of the predicted edges don't correspond to any actual image edges. Therefore, they are not suitable to be SLAM landmark which should be invariant across frames such as point features.
}
\label{fig:other methods not good}
\end{figure}
\fi

%It selects the best room model based on extracted 2D features such as semantics, surface normal, edges and so on.
%A joint volumetric representation with objects and planes can also be directly predicted \cite{tulsiani2017factoring}.

\subsection{Object and Plane SLAM}

For object and plane SLAM, the decoupled approach is to first build classic point SLAM then detect 3D objects and planes \cite{pillai2015monocular}, but it may fail if the point cloud is sparse and not accurate. We here focus on the SLAM which explicitly uses objects and planes as landmarks. Semantic Structure from Motion \cite{bao2012semantic} jointly optimizes various geometry components. Several object based SLAM \cite{salas2013slam++} \cite{galvez2016real} are also proposed but all depend on the prior object models. The recent QuadricSLAM \cite{nicholson2018quadricslam} and CubeSLAM \cite{yang2018object} propose two different object representations for monocular SLAM without prior models. Fusion++ \cite{mccormac2018fusion++} uses RGBD camera to build dense volumetric object models and SLAM optimization.

Concha \cite{concha2015dpptam} utilizes superpixel to provide local planar depth constraints in order to generate a dense map from sparse monocular SLAM. Lee \cite{lee2017joint} estimates the layout plane and point cloud iteratively to reduce mapping drift. Similarly, planes are shown to provide long-range SLAM constraints compared to points in indoor building environments \cite{hsiao2017keyframe} \cite{syang2016popslam}. Recently, \cite{hosseinzadeh2018towards} proposes similar work to jointly optimize objects, planes, points with camera poses. The difference is that we use a monocular camera instead of RGBD camera and also have different object representations.

\section{Single image understanding}
\label{sec:single image}
\begin{figure}[t!]
  \centering
   \includegraphics[scale=0.23]{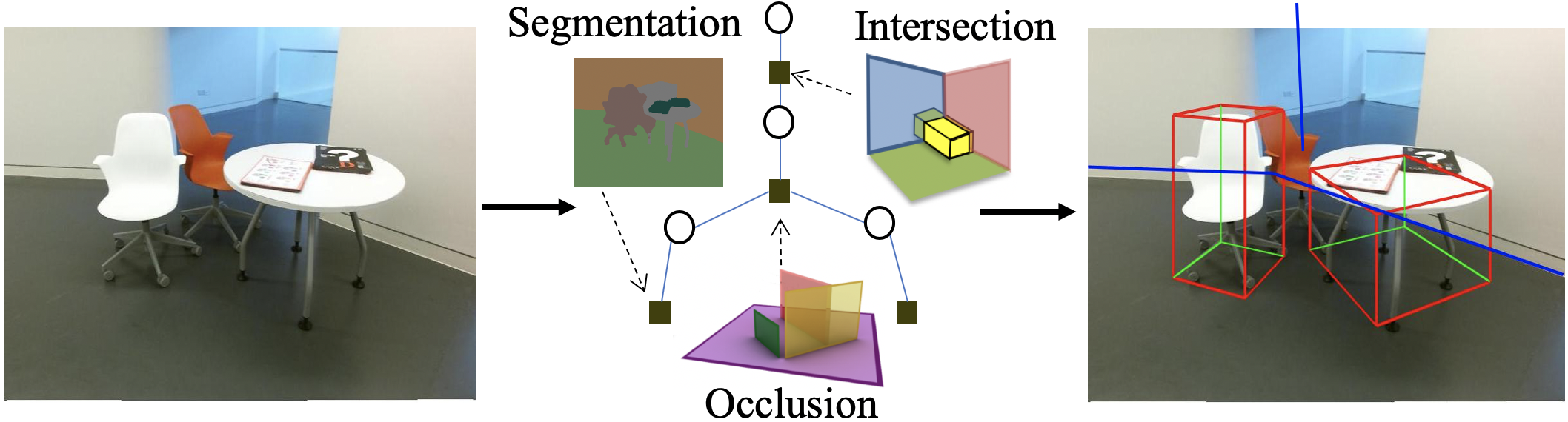}
   \caption[Overview of single image joint 3D object and layout detection]{Overview of single image 3D object and layout detection. We first generate many high-quality object and layout proposals then formulate a graphical model to select the optimal subset based on evidence of semantic segmentation, intersections, occlusions, and so on.
   }
  \label{fig:single image crf illustration}
\end{figure}

% refer to text in proposal.
We represent the environment as a set of cuboid objects and layout planes such as wall and floor. The goal is to simultaneously infer their 3D locations from a 2D image. We first generate a number of object and plane proposals (hypothesis), then select the best subset of them by Conditional Random Field (CRF) optimization, as shown in Fig. \ref{fig:single image crf illustration}.

To represent the layout planes, CNNs can directly predict the 3D plane positions but may lose some details as the predicted layout may not exactly match the actual plane boundary. Therefore the large measurement uncertainty makes it unsuitable to be SLAM landmarks. Instead, we directly detect and select ground-wall line segments which are more reliable and reproducible.

\subsection{Proposal generation}
\label{sec:Proposal generation}

\subsubsection{Layout Plane Proposal}
We first detect all image edges then select some of them close to the ground-wall semantic segmentation \cite{badrinarayanan2017segnet} boundary. For room environments, layout plane prediction score \cite{ren2016coarse} is additionally used to filter out possible edges. If the edge lies partially inside object regions due to occlusions, we further extend it to intersect with other edges.

%We project the actual detected ground-wall edges to 3D space to generate plane proposals, which can be directly used as the latter SLAM landmark because edge observation is consistent across frames. Our prior work \cite{syang2016popslam} also adopts this idea and we extend it to work robustly in large environments with objects.

\iffalse
\begin{figure}[t]
\centering
\subfigure[]{
 \includegraphics[width=1.55in,height=1.1in]{potential_ground.jpg}
 \label{fig:plane proposal sample} 
}
\subfigure[]{
 \includegraphics[width=1.55in,height=1.1in]{multi_3d_object.jpg}
 \label{fig:object proposal sample} 
}
\caption{ Single image generated plane and cuboid proposals. (a) Wall plane proposals are represented as ground edges. (b) Different cuboid proposals for the same object instance.
}
\end{figure}
\fi

\subsubsection{Object Cuboid Proposal}

We follow CubeSLAM \cite{yang2018object} to generate cuboid proposals based on 2D object detection and then score proposals based on image edge features. For each object instance, we select the best 15 cuboid proposals for latter CRF optimization. More cuboid proposals may improve the final performance but also increase computation a lot.

\subsection{CRF Model definition}
\label{sec:CRF model}

Given all the proposals, we want to select the best subset from them. We assign a binary variable $x_i \in \lbrace 0,1 \rbrace$ for each plane and cuboid proposal, indicating whether it will be selected or not. Note that CRF only determines whether it appears or not and doesn't change the proposal's location. The labels are optimized to minimize the following energy function, which is also called potentials in CRF:

\begin{equation}
   \label{eq:total energy}
   \begin{split}
    E(\mathbf{x|I}) = \sum_i \psi^U(x_i)+  \sum_{i<j}  \psi^P(x_i,x_j) + \sum_{\mathbf{x_c}<\mathcal{C}} \psi^{Ho}(\mathbf{x_c})
\end{split}
\end{equation}

\noindent where $\mathbf{x}$ contains all the variables and $I$ is the image. $\psi^U$ and $\psi^P$ are the unary and pairwise potential energy. $\psi^{Ho}(\mathbf{x_c})$ is the high order potential of clique $\mathbf{x_c}$. A clique is a set of variables $x_i$ which have relationships with each other. These potentials are explained in the following:
%Now the CRF inference problem of maximizing $P(\mathbf{x}|\mathbf{D})$ changes to find $\mathbf{x^*} = \arg\min_{\mathbf{x}} E(\mathbf{x}|\mathbf{D})$. 

\subsubsection{Unary potential}
%It represents the proposal quality itself. For planes, the energy depends on the edge's distance to the ground-wall segmentation contour and layout edge prediction score if in room environments. Long edges are also preferred compared to short ones which are likely to be outliers due to detection errors. 
%Proper weighting and normalization is needed to combine them together. 

The unary energy indicates the quality of the proposal. For each proposal, we assign negative unary energy to encourage it. Then in the optimization stage, due to the positive potential from pairwise or high order constraints, only part of them can be selected.

For the wall plane edges, the unary cost is determined based on the semantic segmentation. In more detail, from the segmentation, we can find the ground and wall boundary contour denoted as $c$. Then for each plane proposal edge $x_i$, we sample ten points on the edge and summarize their distance to the contour denoted as $D(x_i,c)$. To compare different edges effectively, we normalize the distance to $d(x_i,c) \in [0,1]$. Then the plane edge unary is defined as:

\begin{equation}
\psi^U(x_i) = -w_i x_i \theta(x_i) \left( 1- d(x_i,c) \right)
\end{equation}

%If $x_i=0$, $\psi^U(x_i)=0$ namely no energy is assigned if the proposal doesn't appear.

\noindent where $w_i$ is the weight for plane unary. $\theta(x_i)$ is the edge's field of view angle to the camera center. A larger angle usually indicates smaller edge detection error. 

For objects, we use the normalized cuboid fitting error explained in \cite{yang2018object}. If the cuboid's edges align better with the detected image edges and vanishing points, the unary will be smaller.

\subsubsection{Pairwise Potential}
%There are different forms of pairwise relationships between objects and planes for example the semantic co-occurrence \cite{lin2013holistic}. Here we only utilize the geometric relationship to minimize the 3D occlusion and intersection. For object-object, $\psi_{ij}^P$ is defined as the 3D intersection of union. For object-plane, it represents the truncation ratio of object volume by plane. For plane-plane, $\psi_{ij}^P$ indicates the angle overlapping ratio between each other. Note that there is no pairwise potential between cuboid proposals belonging to the same object.

There are different forms of pairwise relationship between objects and planes, for example the semantic co-occurrence\cite{lin2013holistic}. Here we only utilize the geometric relationship to minimize the 3D occlusion and intersection.

\begin{figure}[h]
\centering
\subfigure[]{
 \includegraphics[trim={0cm 11.5cm 24.5cm 0cm},clip,scale=0.37]{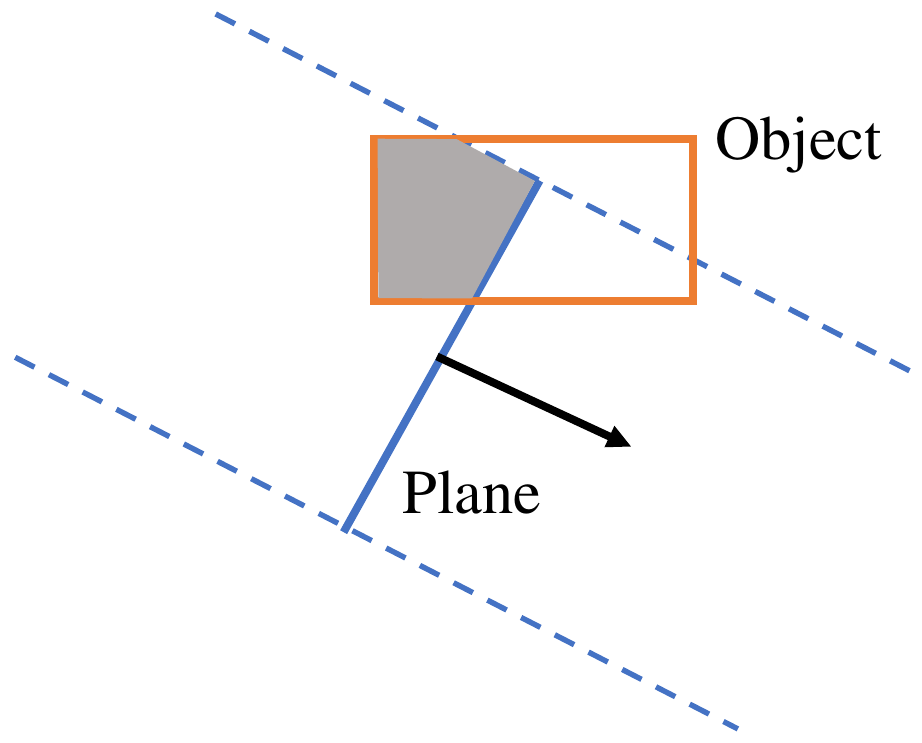}
 \label{fig:object layout}
}\hspace{4em}
\subfigure[]{
 \includegraphics[trim={0cm 12cm 26.5cm 0cm},clip,scale=0.37]{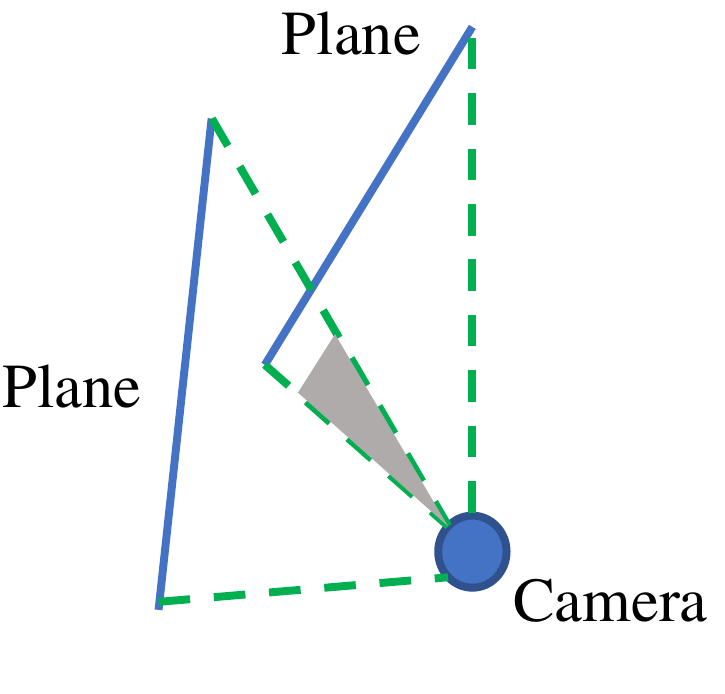}
 \label{fig:layout layout}
}
\caption{Top view of potential definition. (a) Object-plane $\psi_{\mathcal{O}-\mathcal{L}}^P$ potential. The grey part of the object is occluded by the plane. The ratio of grey volume is defined as the potential. (b) Plane-plane $\psi_{\mathcal{L}-\mathcal{L}}^P$ potential. The two planes have angle overlapping and occlusion (grey area) between each other. The potential is defined as the ratio of overlapped angle by total angle.
}
\end{figure}

Object-object potential $\psi_{\mathcal{O} -\mathcal{O}}^P$ is defined as the 3D intersection of union between two cuboids as in Eq (\ref{eq:obj obj pair}). In the equation, if two proposals both appear, namely $x_i,x_j=1$, the potential becomes positive otherwise it is zero. Note that there is no pairwise potential between cuboid proposals belonging to the same object instance. Object-plane $\psi_{\mathcal{O}-\mathcal{L}}^P$ potential depends on the volume ratio of the object occluded by plane shown in Eq (\ref{eq:obj plane pair}) and Fig. \ref{fig:object layout}. Similarly, plane-plane $\psi_{\mathcal{L}-\mathcal{L}}^P$ is defined as the angle overlapping ratio between each other in Eq (\ref{eq:plane plane pair}) in Fig. \ref{fig:layout layout}. Since large plane occlusion is strongly discouraged, an infinite penalty cost is assigned if their overlapping angle is greater than 5$\degree$.

\begin{equation}
\psi_{\mathcal{O}-\mathcal{O}}^P(x_i,x_j) = x_i x_j \frac{V(x_i)\bigcap V(x_j)}{V(x_i)\bigcup V(x_j)}
\label{eq:obj obj pair}
\end{equation}

\begin{equation}
\psi_{\mathcal{O}-\mathcal{L}}^P(x_i,x_j) = x_i x_j \frac{V^{occ}(x_i)}{V(x_i)}
\label{eq:obj plane pair}
\end{equation}

\begin{equation}
\psi_{\mathcal{L}-\mathcal{L}}^P(x_i,x_j) = x_i x_j \frac{A(x_i)\bigcap A(x_j)}{A(x_i)\bigcup A(x_j)}
\label{eq:plane plane pair}
\end{equation}

\noindent Where $V(x)$ denotes the 3D object volume and $A(x)$ denotes plane angle range to camera center. $\bigcap$ represents intersection and $\bigcup$ is union.

\subsubsection{High order potential}
\label{sec:high order poten}
As explained in Section \ref{sec:Proposal generation}, for each 2D object instance, many 3D cuboid proposals are generated from it but at most one of them can be selected. Thus these 3D proposals from one object form a clique $\mathbf{x_c}$ and the high order potential is defined as:

\begin{equation}
\label{eq:high order eq}
\psi^{Ho}(\mathbf{x_c})=\left\{
\begin{aligned}
   & 0  \quad \;\,  \text{if} \ \sum_{x_i \in \mathbf{x_c}} x_i\leq 1   \\
   & \infty \quad \text{otherwise}
\end{aligned}
\right.
\end{equation}

\subsection{Efficient CRF inference}
\label{sec:CRF inference}
Efficient inference of high order discrete CRFs is still a challenging problem \cite{wang2013markov}. We observe that the high order term in Eq \ref{eq:high order eq} is very sparse because at most one variable can be 1 in one clique $\mathbf{x_c}$. We therefore design efficient inference based on max-product loopy belief propagation \cite{koller2009probabilistic}. After the iterative message passing, we select the state with minimum potential as the final result. The computationally expensive part is the message from clique $c$ to variable node $i$:

\begin{align}
&m_{c\rightarrow i}^t (x_i) = \min_{\mathbf{x}_c^{-i}} \bigg( f_c(\mathbf{x}_c)+\sum_{j\in c \setminus \lbrace i \rbrace } m_{j\rightarrow c}^{t-1}(x_j) \bigg) \label{eq:factor to variable}
\end{align}

\noindent where $\mathbf{x}_c^{-i}$ denotes all the variables in clique $c$ except variable $i$. $t$ and $t-1$ represents different iteration steps. $m_{j\rightarrow c}^{t-1}(x_j)$ is the message from node to clique.  For a clique with $N$ binary nodes, there are totally $2^N$ clique states of $\mathbf{x}_c$. However there are only $N+1$ valid states in our problem $\lbrace1,0,...0\rbrace,...\lbrace0,0,...1\rbrace,\lbrace 0,0,...0 \rbrace$ denoted as $\big \lbrace \mathbf{y}_1,\mathbf{y}_2,...\mathbf{y}_{N+1}   \big \rbrace$. Therefore, we only need to check $N+1$ states and find the minimum in Eq \ref{eq:factor to variable}. We can further observe that every adjacent state vector $\mathbf{y}_i$ only has two different variables, therefore $\sum_{j\in c \setminus \lbrace i \rbrace }m_{j\rightarrow c}^{t-1}(x_j)$ for each $\mathbf{y}_i$ can be computed iteratively. The average time complexity of computing $m_{c\rightarrow i}^t (x_i)$ is $O(1)$ instead of the naive $O(2^N)$. More details can be found at the appendix.

\section{SLAM Optimization}
\label{sec:slam}
The selected object and plane proposals from single image detection are used as SLAM landmarks and optimized together with camera poses through multi-view BA. We also include points in SLAM because there are usually only a few objects and planes in the environments and they cannot fully constrain camera poses. In the following, we first formulate the optimization problem, then explain the parameterization and various measurement costs.

\subsection{Bundle Adjustment Formulation}
\label{sec:slam pipeline}
Consider a set of camera poses $C=\{c_i\}$, 3D objects $O=\{o_j\}$, planes $\Pi = \{\pmb{\pi}_k \}$ and points $P=\{p_m\}$, bundle adjustment can be formulated as nonlinear least squares optimization problem:

\begin{equation}
C^*,O^*,\Pi^*,P^* = \arg \min_{\{C,O,\Pi, P\}} \sum_{i\in C, j \in O, k \in \Pi, m \in P} \mathbf{e^T} \Sigma \mathbf{e}
\label{eq:general obj ba}
\end{equation}

\noindent where $\mathbf{e}$ is the measurement error between each other.  $\Sigma$ is covariance matrix of different error measurements. The optimization problem can be solved by Gauss-newton or Levenberg-Marquardt algorithm in many libraries such as g2o and iSAM.

%\vspace{0.5em}
%\noindent \textit{Notations}: Camera poses are represented by $T_c \in SE(3)$ and points are represented by $P \in \mathbb{R}^3$. As explained in Section \ref{sec:obj detect core}, cuboid objects are modelled as 9 DoF parameters: $O=\{T_o, \mathbf{d}\}$ where $T_o =[R \ \mathbf{t}] \in SE(3)$ is 6 DoF pose, and $\mathbf{d} \in \mathbb{R}^3$ is the cuboid dimension. In some environments such as KITTI, we can also use the provided object dimension then $\mathbf{d}$ is not needed to optimize. Subscript $m$ indicates the measurement. The coordinate system is shown in Fig \ref{fig:slam measurement}.

\subsection{Parameterization}
\label{sec:slam parameterization}

For camera pose and point, we utilize the standard form $T_c \in SE(3)$ and $P \in \mathbb{R}^3$. The cuboid pose is similarly defined in \cite{yang2018object} by 9 DoF parameters: $O = (T_o, D)$, where $T_o \in SE(3)$ is 3D object pose, and $D \in \mathbb{R}^3$ is dimensions.

We adopt the infinite plane representation \cite{kaess2015simultaneous} $\pmb{\pi}=(\mathbf{n}^\top,d)^\top$  st. $\|\pmb{\pi}\|=1$. $\mathbf{n}$ is the plane normal and $d$ is the plane distance to the world origin. In some environments, we use the Manhattan assumptions, namely the plane normal is fixed and parallel to one of the world frame axes, therefore only $d$ is needed to represent it.

\subsection{Measurements}
\label{sec:slam meas}

Different measurement functions between the map components are proposed to formulate factor graph optimization. Camera-point observation model is the standard point reprojection error \cite{mur2015orb}. We here explain the new measurements in more detail.

\subsubsection{Camera-plane}
\begin{figure}[t]
\centering
\vspace{0.55em}
\subfigure[]{
 \includegraphics[width=2.5in,height=0.8in]{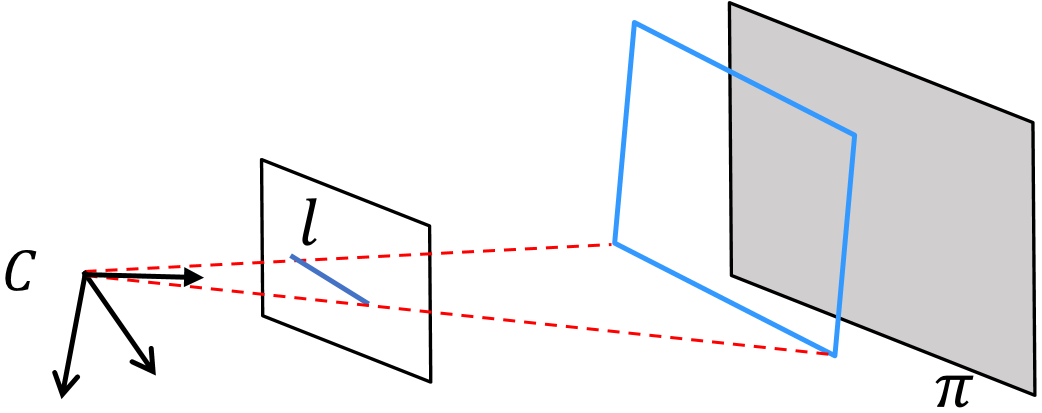}
 \label{fig:camera plane} 
}
\subfigure[]{
 \includegraphics[width=2.2in,height=0.7in]{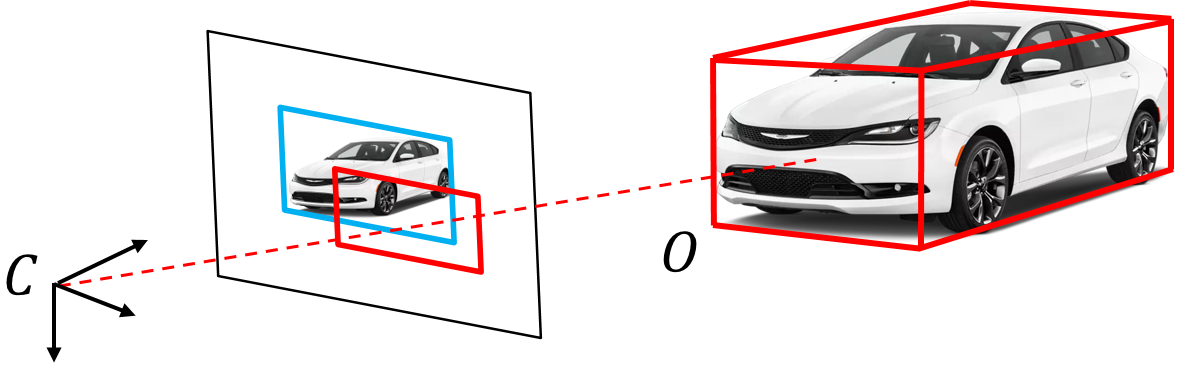}
 \label{fig:camera object} 
}
\hspace{0.5em}
\subfigure[]{
 \includegraphics[width=0.8in,height=0.7in]{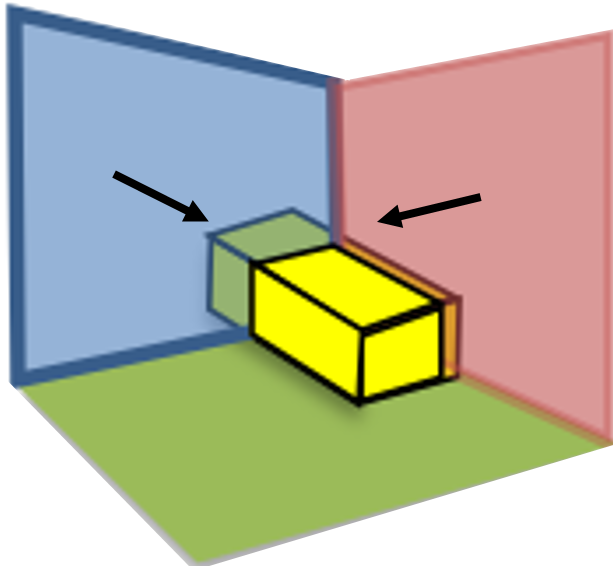}
 \label{fig:object plane} 
}
\caption{ SLAM observation functions. (a) Camera-plane observations. The detected ground edge is back-projected to 3D space to compare with landmark plane. (b) Camera-object observations. 3D cuboid landmark is projected onto images and compared with the detected 2D box. (c) Object-plane measurement error depends on the object volume occluded by planes.
}
\end{figure}

Different from RGBD based plane SLAM which can directly get plane measurement from point cloud plane fitting \cite{hosseinzadeh2018towards} \cite{kaess2015simultaneous}, we need to back-project 2D plane edge $l$ to the 3D space to get the measurement shown as the blue plane in Fig. \ref{fig:camera plane}, then compare it with the grey plane landmark plane $\pmb{\pi}$ using $\log$ quaternion error:

\begin{equation}
e_{cp} = \| \log \big( \pmb{\pi}_{obs}(l), T_c^T \pmb{\pi} \big) \|
\label{eq:plane error}
\end{equation}

%$\pmb{\pi}_{obs}(l)$ 

Note that in Eq \ref{eq:plane error}, we transform the global plane landmark to camera frame by $T_c^T \pmb{\pi}$ instead of comparing them in the world frame. This is because when camera moves far away from world origin, plane parameter $d$ becomes very large compared to normal $\mathbf{n}$ and dominate the error.

For the back-projection process, suppose $K$ is the camera intrinsic calibration and $p$ is one of the endpoints of edge $l$, then its corresponding 3D point $P$ is the intersection of back projected ray $K^{-1}p$ with the ground plane $(\mathbf{n}_g^\top,d_g)$ in camera frame:
\begin{equation}
\label{eq:projection equation}
P=\frac{-d_g}{\mathbf{n}_d^\top(K^{-1}p)}K^{-1}p
\end{equation}

Similarly we can compute the other endpoint and get the 3D vertical wall plane $\pmb{\pi}_{obs}$ passing through the two points. We can find that this process depends on the camera pose to the ground plane. Therefore, we need to update it each iteration during the optimization. 

%\cite{syang2016popslam} updates the measurement after graph optimization which is not an optimal solution. 

\subsubsection{Camera-object}
We follow the cuboid observation functions defined in the prior work \cite{yang2018object}. The cuboid landmark's 3D corners are first projected onto the image plane then a 2D rectangle hull is found shown as the red rectangle in Fig. \ref{fig:camera object}. Then it is compared with the blue actual detected 2D bounding box:

\begin{equation}
\label{eq:object camera 2d}
e_{2D} = \parallel [\mathbf{c}, \mathbf{s}]-[\mathbf{c_m},\mathbf{s_m}]\parallel_2
\end{equation}

\noindent where $[\mathbf{c}, \mathbf{s}]$ is the center and dimension of the 2D box. This 2D measurement error has much less uncertainty compared to 3D cuboid error as explained in \cite{yang2018object}. To make the optimization robust, we assign different weights to different objects based on their distance to camera and 2D semantic object detection confidence.

\subsubsection{Object-plane}
There are different forms of object-plane constraints depending on the environment assumptions for example objects are supported by planes \cite{hosseinzadeh2018towards} or object orientation matches the nearby plane normal. We here propose a weaker but more general constraint that objects should not be occluded by nearby planes in the camera view shown in Fig. \ref{fig:object plane}. The error is the sum of 3D corners' signed distance to plane:

\begin{equation}
e_{op} = \sum_{i=1:8}  \max(0,-\pmb{\pi} P_{oi})
\end{equation}

\noindent where $P_{oi}$ is one of the eight cuboid corners. If the cuboid lies on the positive side of the plane meaning that there is no occlusion, $e_{op}$ will be zero.

%The plane normal is defined towards the camera, then

\subsubsection{Point-plane}
\label{sec: point plane error}
%If point is detected to belong to the plane region, we can also add a straightforward geometric constraint. However,
If a feature point belongs to a plane region, we also add a constraint of the point's 3D distance to plane. However, it is usually difficult to accurately determine if a point belongs to a plane from 2D image as layout planes are usually the background and points may belong to the foreground objects. To improve the robustness, we first select feature points in the 2D wall plane polygon then filter out points that are farther away from the 3D plane than a threshold. The point-plane error is defined as:

\begin{equation}
e_{pp} = \| \pmb{\pi} P \|_2
\label{eq:point plane}
\end{equation}

\noindent Note that to be robust to outliers, huber loss is applied to all above error functions.

\subsection{Data association}

Data association for different landmarks across multiple views is necessary to build a SLAM graph. For point association, we use the point feature matching in ORB SLAM \cite{mur2015orb}. Object association follows the work of CubeSLAM \cite{yang2018object}. Basically, each object contains a set of feature points belonging to it, then we can find object matching which has the most number of shared map points in different views. This approach is easy to implement and can effectively handle occlusion, repetitive textures and dynamic movement.

For the plane association, we first check whether the plane normal difference is within $30 \degree$ and distance to each other is smaller than $1m$. We then find the plane matching with the most shared feature points similar to object matching. In Sec \ref{sec: point plane error}, we already determine which feature points belong to the specific plane.

\section{Experiments}
\label{sec:experiment}
\subsection{Implementation details}

For object detection, we use similar settings as object SLAM in \cite{yang2018object}. Yolo \cite{redmon2016yolo9000} detector is used for the 2D object detection. For plane detection, we first detect line segments using \cite{von2008lsd} and merge them to long edges. Segnet \cite{badrinarayanan2017segnet} is used for the 2D semantic segmentation. We then filter out lines whose length is shorter than 50 pixels and more than 50 pixels away from the wall-ground segmentation boundary. In video case, SLAM pose estimation is used in the single image 3D detections.

%Having many false positive wall planes will increase difficulty for the latter CRF selection and latter SLAM association and optimization.

%The explicit image recognition based loop closure detection in ORB SLAM is disabled in order to better show the improvements on the pose estimation by objects and planes.

For the SLAM part, our system is built on the feature point based ORB SLAM, augmented with our objects and planes. We compute jacobians of new observation functions then perform BA using g2o library. Since the outlier associations and measurements of objects and planes have more severe effects on the optimization compared to outlier points, strict outlier rejections have to be utilized. In our system, the object and plane landmark will be deleted if it has not been observed by 3 frames in recent 15 frames after creation or if there are less than 10 stable feature points associated with it. In most of the room environments, we use the Manhattan plane representation with a fixed surface normal as mentioned in Section \ref{sec:slam parameterization} to provide more constraints on the planes and improve the overall performance. If the initially generated wall surface normal difference with Manhattan direction exceeds 30 degrees, it will also be treated as an outlier.

%except the white wall surfaces with few 2D features initially

In addition to being used as SLAM landmarks, objects and planes also provide depth initialization for those feature points which are difficult to triangulate due to small motion parallax. When less than 30\% of the feature points are matched to map points, we create some new map points directly using depth from objects and planes. This can improve monocular SLAM performance in low texture environments and large rotation scenarios. Compared to the prior work of monocular plane SLAM \cite{syang2016popslam}, ground plane is not used in this work because there is no actual edge measurement corresponding to the ground plane. 

For the final dense map generation, we back-project pixels in the plane regions onto the optimized plane landmarks. For feature points belonging to objects, we create triangular meshes in 3D space to get a dense 3D model. Note that in the SLAM optimization, planes are represented as infinite planes, but for visualization purposes, we also keep track of the plane boundary polygon.
% in order to back project its contained pixels

\subsection{Single Image Result}

\begin{figure}[t]
\centering
\vspace{0.25em}
\includegraphics[width=1.1in,height=0.8in]{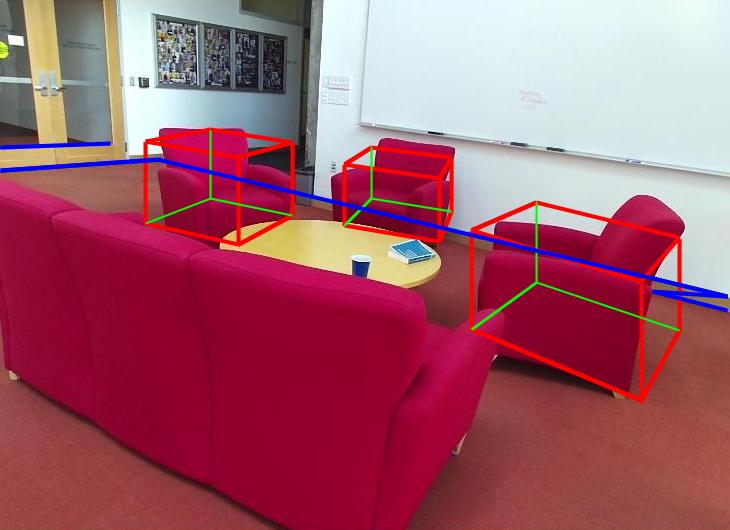}
\hspace{0.05cm} \includegraphics[width=0.98in,height=0.8in]{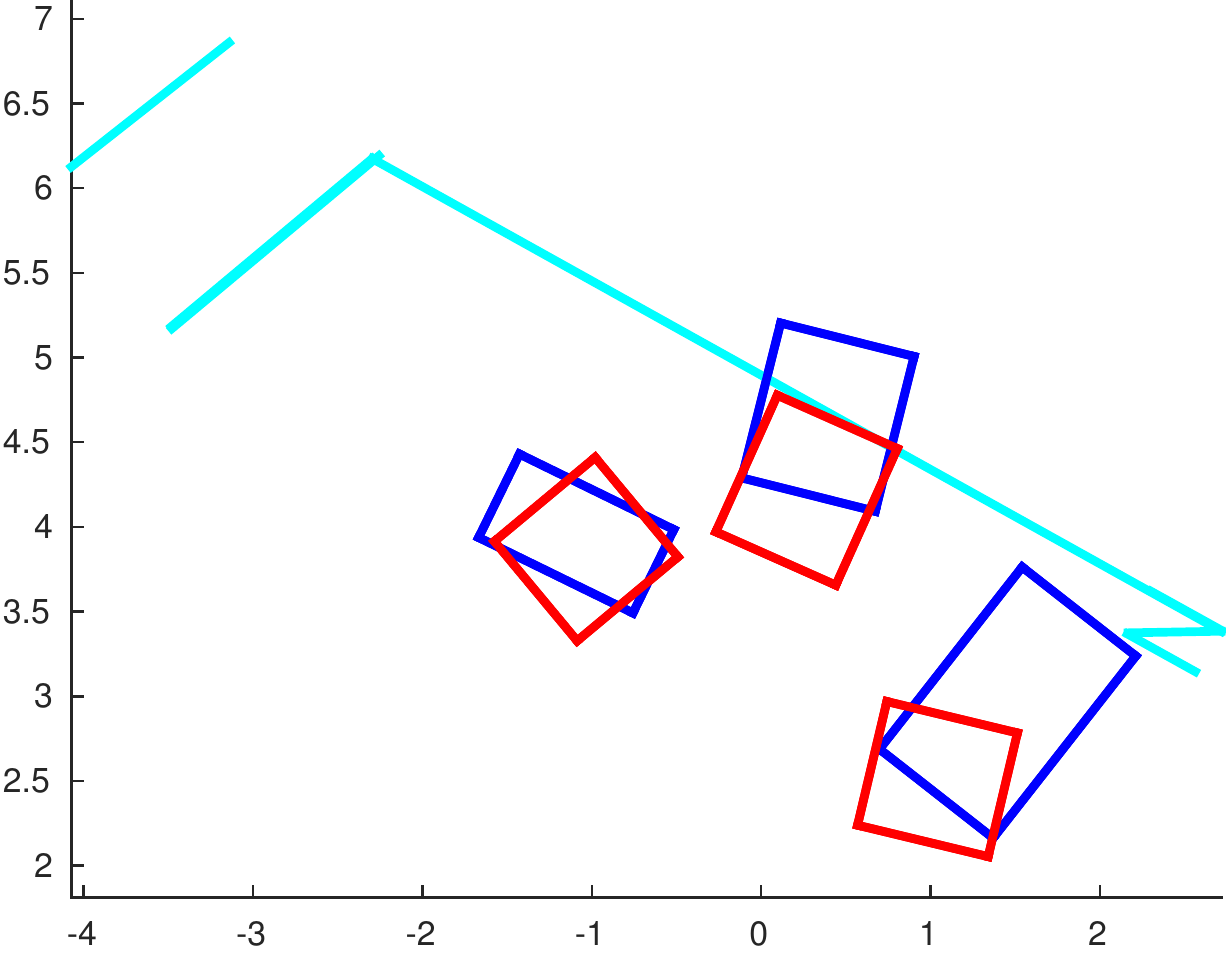}
\hspace{0.05cm} \includegraphics[width=0.98in,height=0.8in]{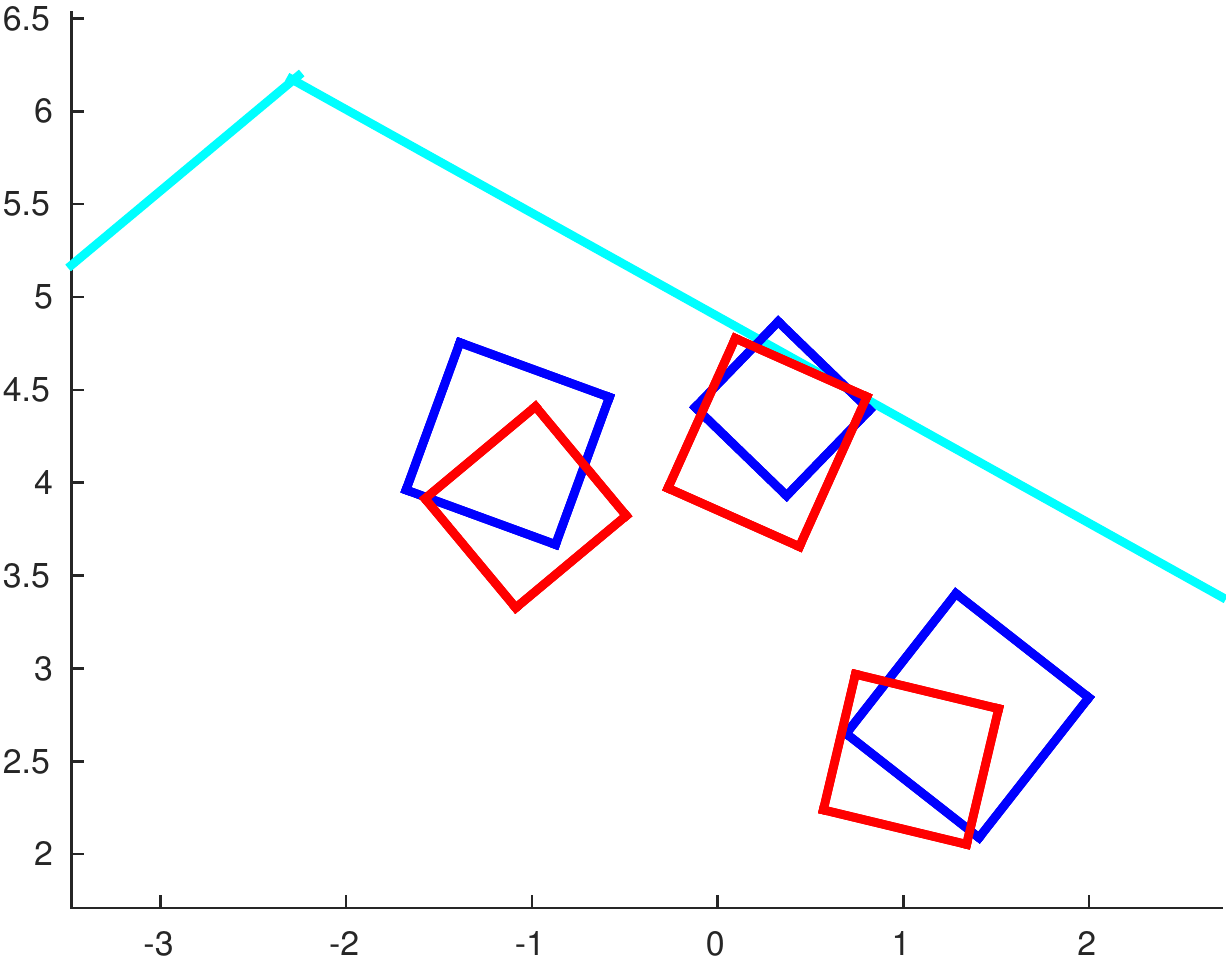}
\subfigure[]{\hspace{0.01cm} \includegraphics[width=1.1in,height=0.8in]{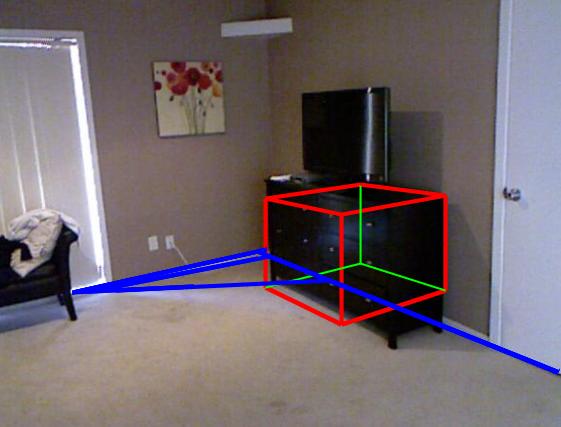}}
\vspace{0.02cm}
\subfigure[]{ \hspace{0.03cm}\includegraphics[width=0.98in,height=0.8in]{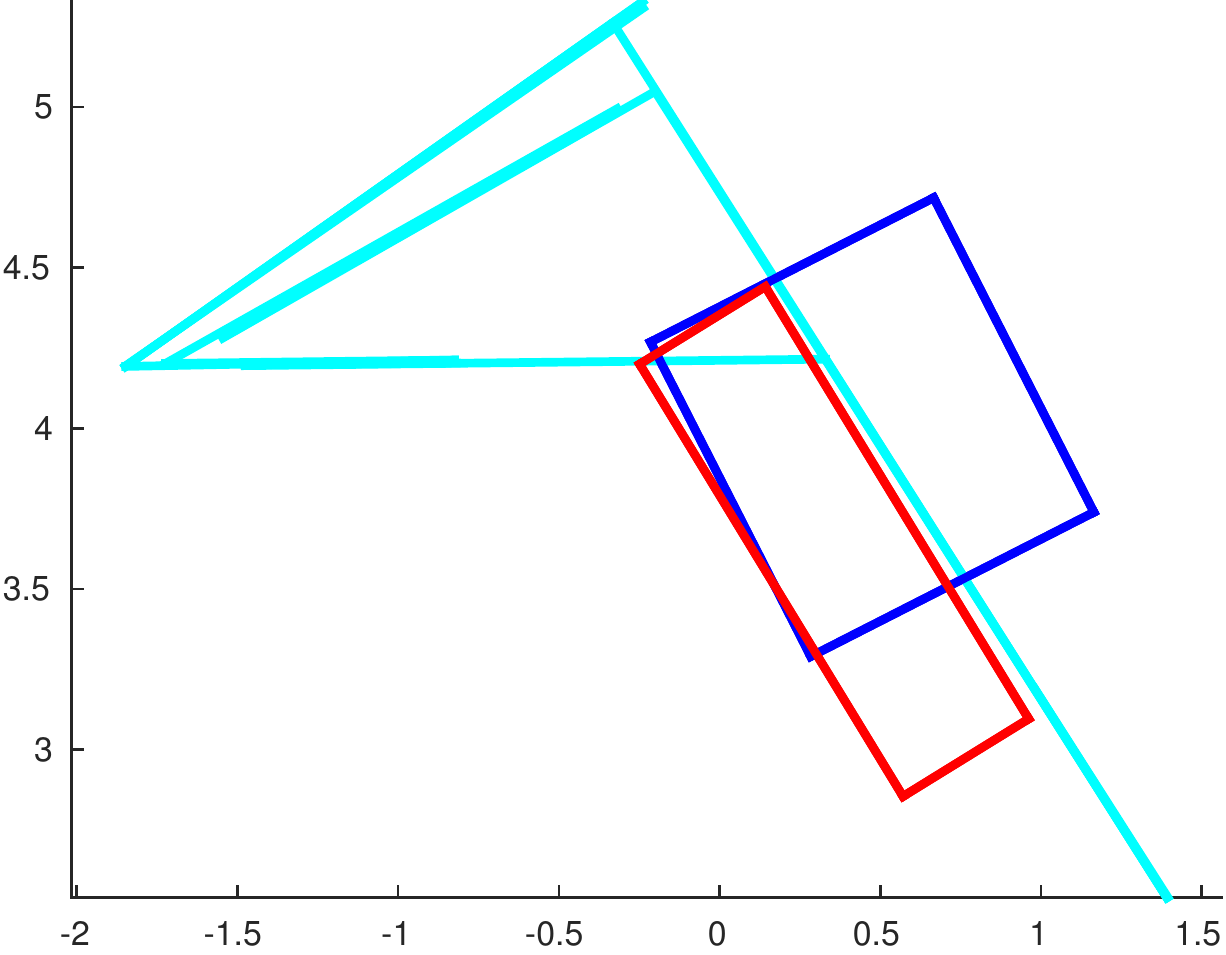} }
\subfigure[]{\hspace{-0.02cm} \includegraphics[width=0.98in,height=0.8in]{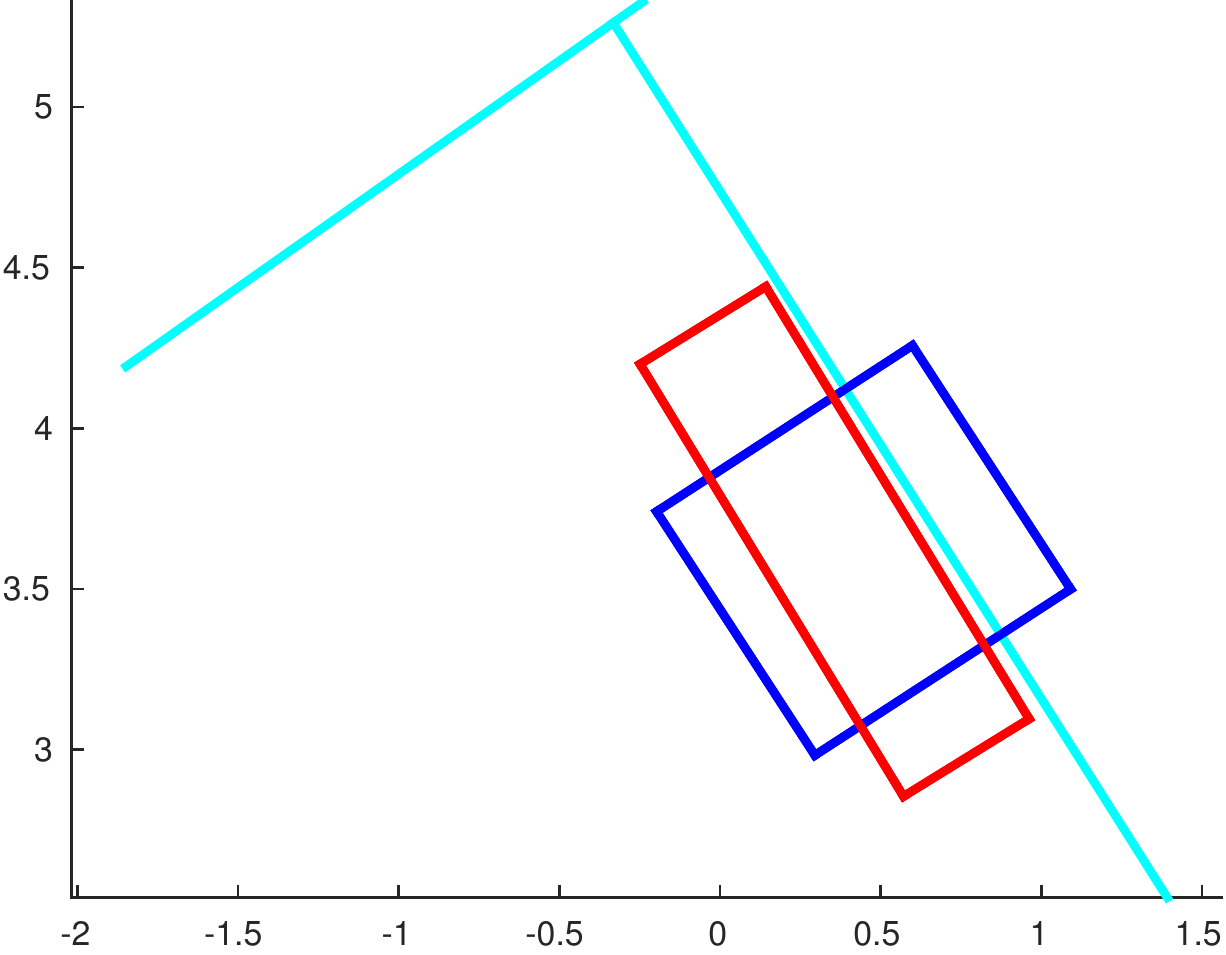} }
\caption{ Single image raw proposal generation and CRF optimization illustrations. (a) Raw plane and object proposals. (only one cuboid proposal is drawn for brevity) (b) Top view of raw proposals. The red rectangle is ground truth objects and blue are the estimated. Cyan lines are wall plane proposals. (c) Top view of CRF selected proposals. Object poses are more accurate after optimization. Plane and object intersection and occlusion is also reduced.
}
 \label{fig:crf opti}
\end{figure}

We first show the single image object and plane result. Some examples of proposal generation and CRF optimization are shown in Fig. \ref{fig:crf opti}. The middle and right columns show the top view of object proposals before and after CRF optimization. We can see that CRF can select non-overlapped wall edges and better cuboid proposals to reduce occlusion and intersection. Since CRF only selects the proposals without changing their actual locations, there might still be occlusion after optimization.

\begin{figure}[t]
\centering
 \includegraphics[width=1.06in,height=0.8in]{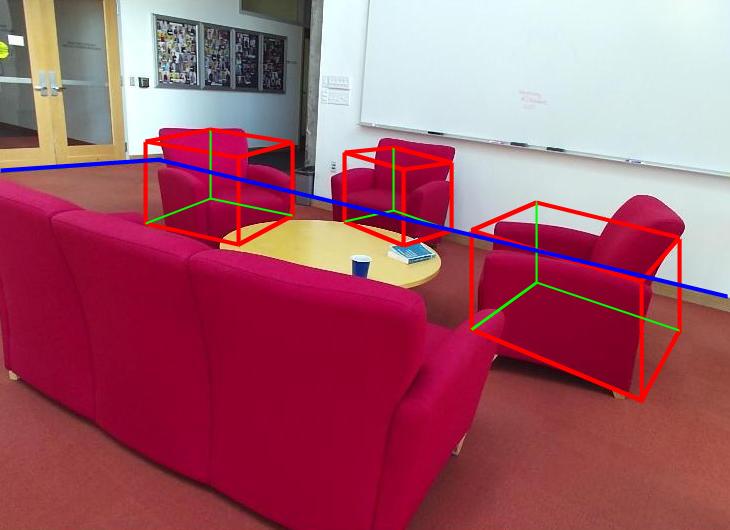}
 \includegraphics[width=1.06in,height=0.8in]{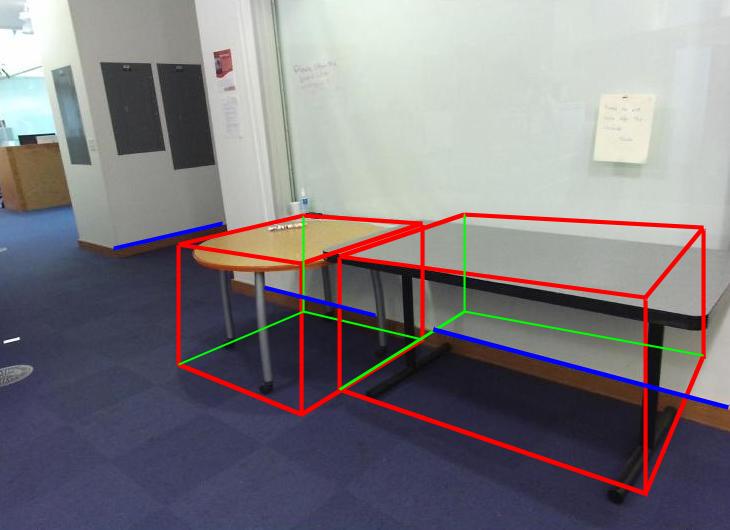}
 \includegraphics[width=1.06in,height=0.8in]{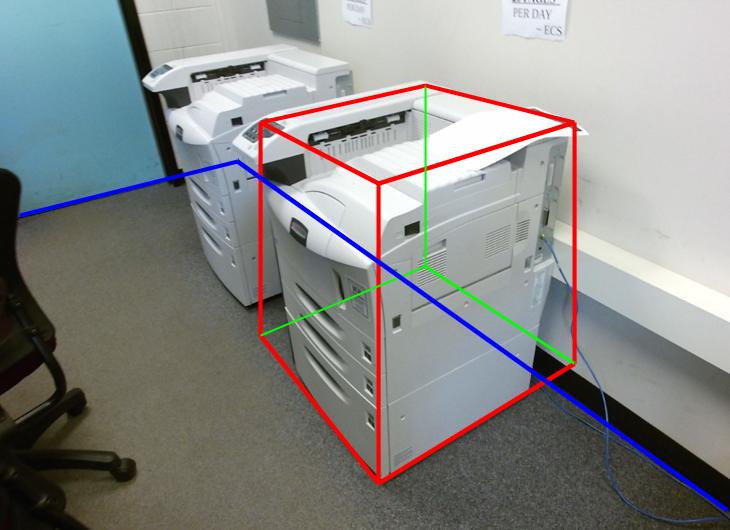}
 \includegraphics[width=1.06in,height=0.8in]{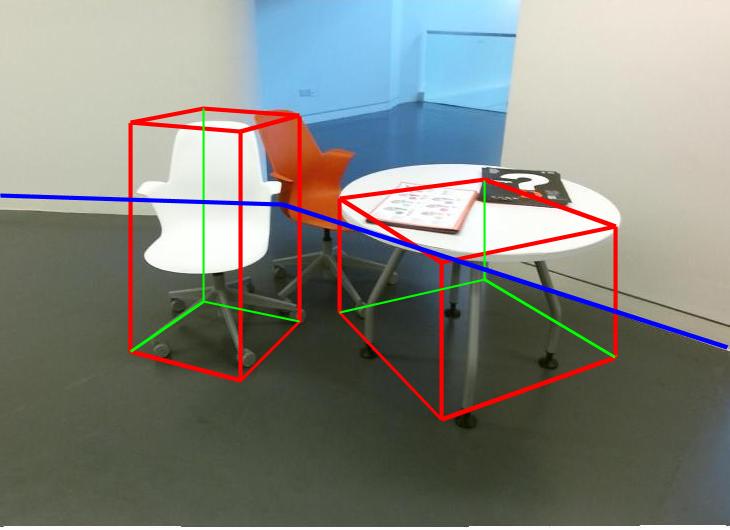}
 \includegraphics[width=1.06in,height=0.8in]{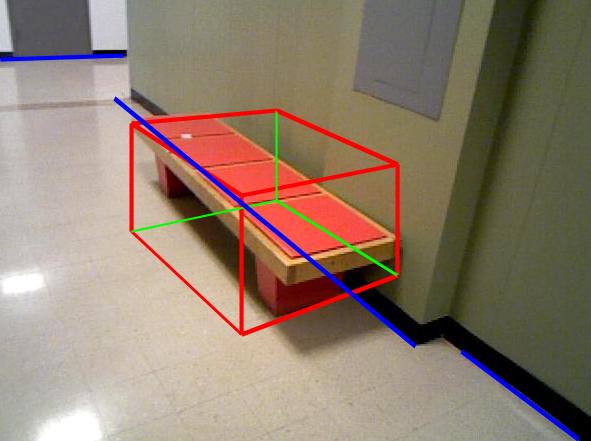} %\hspace{0.05cm}
 \includegraphics[width=1.06in,height=0.8in]{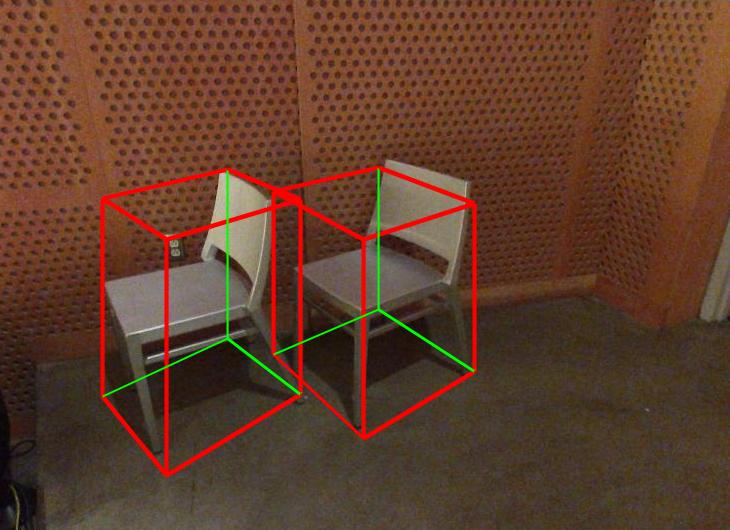} 
\caption{More single image 3D detection results, where blue line represents the wall plane edge and the box represents the object.
}
 \label{fig:single img final}
\end{figure}

More results of CRF selected object and plane proposals are shown in Fig. \ref{fig:single img final}. The algorithm is able to work in different environments including rooms and corridors but it may still miss some planes and objects when there is severe object occlusion and unclear edges for example in the right column of Fig. \ref{fig:single img final}.

We also evaluate quantitatively on the SUN RGBD dataset using the 3D object intersection over union (IoU) as the metric. We select 1670 images with visible ground planes and ground objects fully in the field of view, then compare with the prior work \cite{yang2018object}, the latest deep network based scene understanding \cite{huang2018cooperative} and two other model-based algorithms: SUN primitive \cite{xiao2012localizing} and 3D Geometric Phrases (3dgp)\cite{choi2013understanding}. Note that ground truth camera pose is used in our method as well as \cite{xiao2012localizing} \cite{choi2013understanding}. Huang \textit{et al.}\cite{huang2018cooperative} predict camera pose and layouts jointly so it is difficult to modify their algorithm to use the provided camera pose. From Table \ref{table:single obj iou}, our prior work \cite{yang2018object} performs similar to other work. \cite{huang2018cooperative} performs worse because it utilizes the predicted camera pose therefore it is not quite comparable. Compared to \cite{xiao2012localizing}\cite{choi2013understanding}, our method detects much more objects. The proposed CRF joint optimization improves the IoU by 5$\%$ compared to \cite{yang2018object}. Note that to emphasize the optimization effect, we only evaluate on images where CRF generates different results compared to the single image detection. This is because in most images, there are no wall visible planes or planes are far from objects, therefore planes have no constraints on object positions.

\begin{table}[t]
\vspace{0.25em}
\caption{3D Object IoU on SUN RGBD Subset Data}
\begin{center}
\begin{tabular}{c c c c c c}
\toprule
Method  &Huang\cite{huang2018cooperative}   & Xiao \cite{xiao2012localizing}  & 3dgp \cite{choi2013understanding}   & Ours \cite{yang2018object}  	& Our CRF  \\  \hline
3D IoU 	& 0.27   & 0.30  & 0.35       & 0.38	 		        & 0.43 \\
\bottomrule
\end{tabular}
\end{center}
\label{table:single obj iou}
\end{table}

\subsection{SLAM Result}
We then evaluate the SLAM tracking and mapping performance on both public datasets ICL-NUIM \cite{handa:etal:ICRA2014}, TAMU Indoor \cite{lu2015robustness}, TUM mono \cite{engel2016monodataset}, and our collected datasets by KinectV2 sensor. 

%We select most of the room and corridor sequence where there is not severe object occlusions so that our algorithm can find wall planes.

\subsubsection{Qualitative Results}
\begin{figure}[t]
  \centering
  \subfigure[]{
   \includegraphics[scale=0.12]{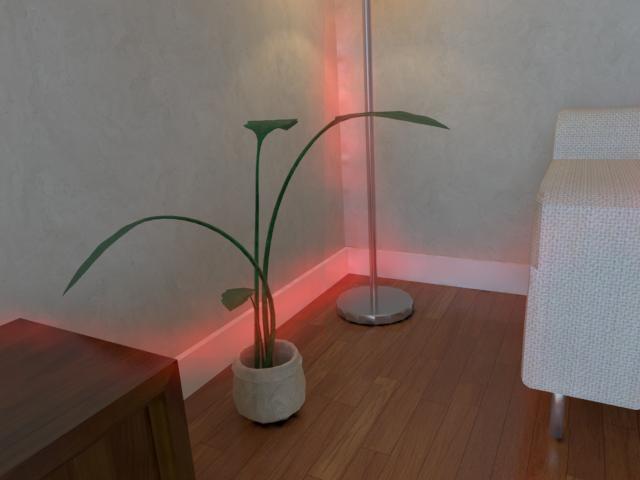}
   \label{fig:icl layout vis}
   }
   \hspace{-1em}
  \subfigure[]{   
   \includegraphics[scale=0.12]{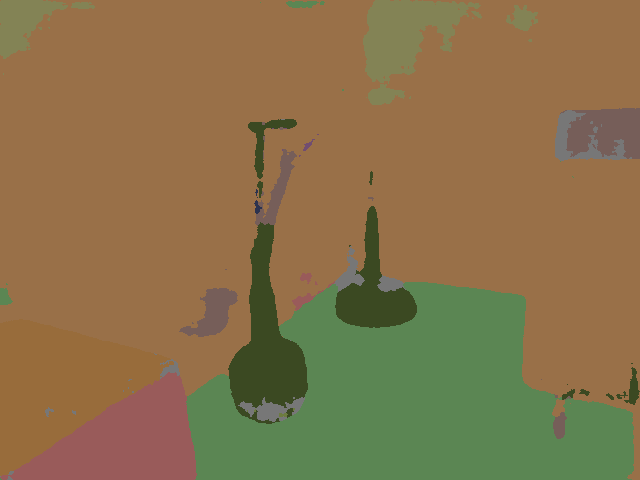}
   \label{fig:icl semantic vis}   
   }
   \hspace{-1em}   
  \subfigure[]{   
   \includegraphics[scale=0.12]{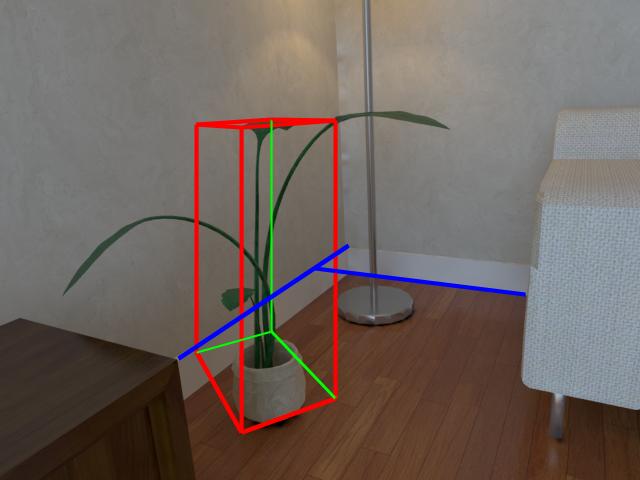}
   \label{fig:icl object layout vis}
   }
   \caption{(a) Layout prediction score map \cite{ren2016coarse} (b) Semantic segmentation by \cite{badrinarayanan2017segnet}
   (c) Our single image object and plane detections. It cannot detect the occluded wall surface while multi-view SLAM can build a complete map in Fig. \ref{fig:intro figure}, demonstrating the advantage of our multi-view object SLAM.
   }
   \label{fig:icl frame}
\end{figure}

A sample frame of ICL sequence is shown in Fig. \ref{fig:icl frame}. The left and middle images show the raw image overlayed by layout prediction and the semantic segmentation. Both of them have noise and our CRF optimization in Fig. \ref{fig:icl object layout vis} shows a roughly correct 3D model but it cannot fully detect the occluded wall segments. After the multi-view SLAM optimization, the algorithm is able to build a more consistent and complete map shown in Fig. \ref{fig:intro figure}.

\begin{figure}[t]
\vspace{0.55em}
  \centering
   \includegraphics[scale=0.15]{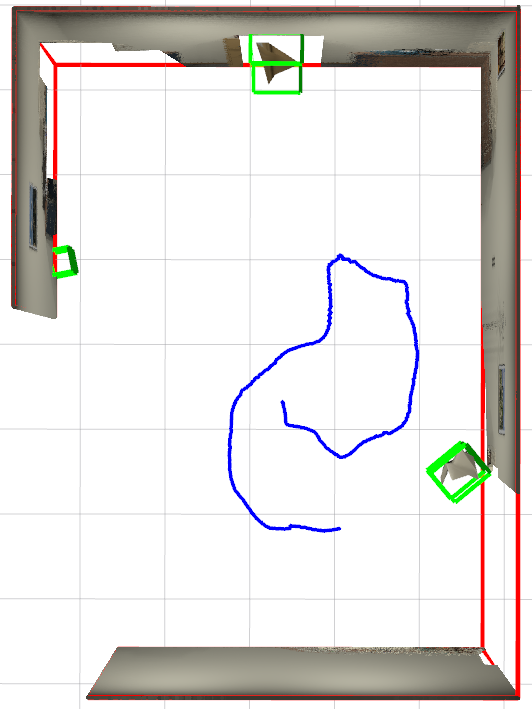} \vspace{0.2cm}
   \includegraphics[scale=0.15]{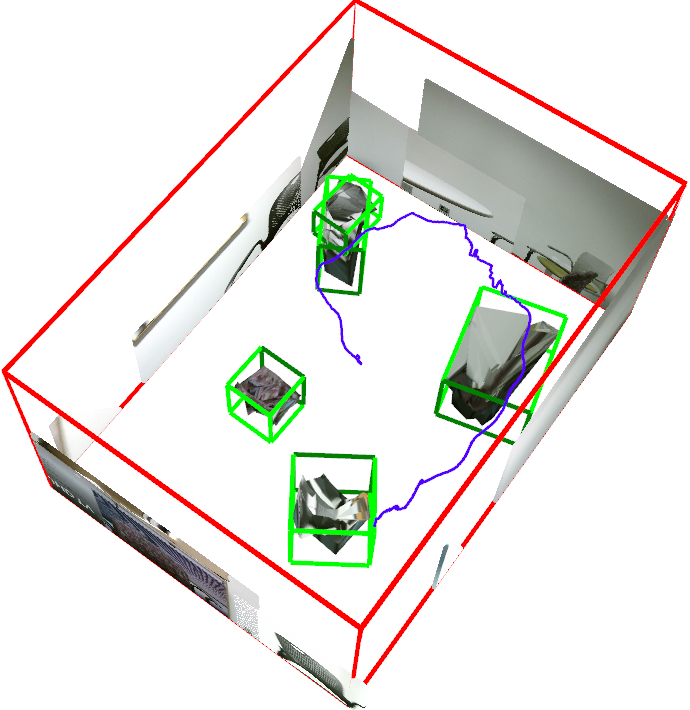} \vspace{0.2cm}
   \includegraphics[scale=0.22]{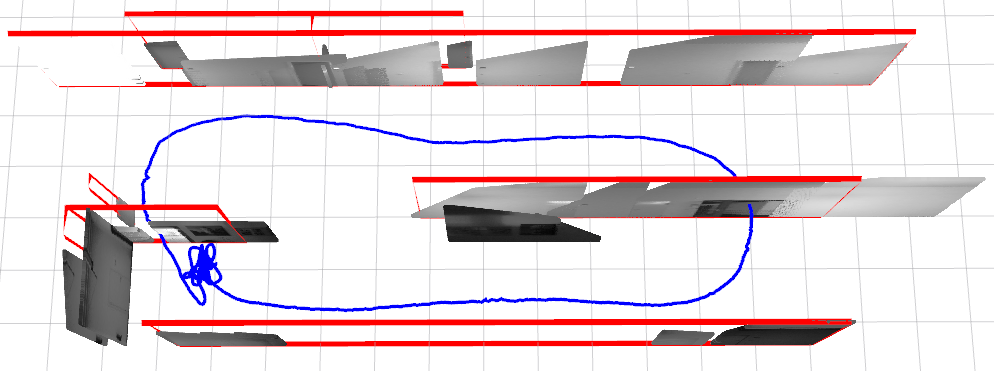}
   \includegraphics[scale=0.19]{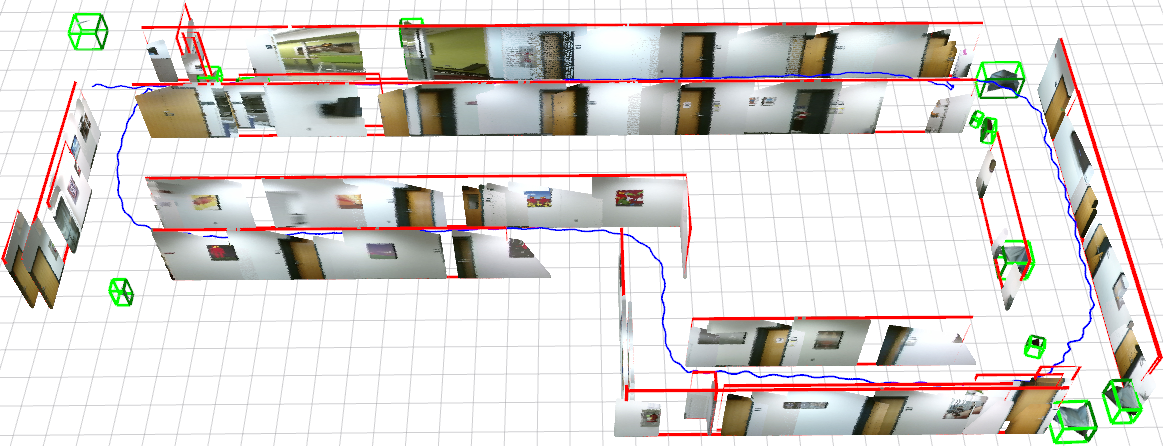}
   \caption{ More dense mapping results with objects and planes. (top) ICL-NUIM office 2 and collected room data. (middle) TUM-mono 36.  (bottom) Our collected long corridor. The red rectangle is the plane boundary and green cuboid is the object. The blue curve is the estimated camera trajectory.}
   \label{fig:more 3d map}
\end{figure}

More examples of 3D mapping and camera pose estimation in different datasets and environment configurations are shown in Fig. \ref{fig:more 3d map}. The green box is the object location and red rectangle is the plane boundary. After BA, objects and planes' locations are more accurate compared to the single view detection and most objects lie inside the room. Note that not all objects are mapped because the 2D object detector might miss some and SLAM might also classify some of them as outliers due to inconsistent observations. In some scenarios such as the top left of Fig. \ref{fig:more 3d map}, our algorithm cannot detect the complete wall plane due to severe object occlusions. To improve the visualization robustness, if there are not enough map points observed in some region of a plane polygon, the pixels won't be back-projected to generate dense maps, shown as the missing segments on the wall surface.

%The top left and middle row is from the public dataset.

\subsubsection{Quantitative Results}

%and generated plane depth
We then show the quantitative camera pose comparison with ORB SLAM and DSO. For datasets in Table \ref{table:ate pose error}, the initial maps of both ORB SLAM and ours are scaled by the ground truth initial camera height. Then we can directly evaluate the absolute translation error without aligning the pose in scale, to show that object and planes can improve the pose estimation and reduce monocular drift. Each algorithm runs 5 times in each sequence and the mean error is reported here. From the table, we can see that in most of the scenarios, the added object and plane landmark constraints improve the camera pose estimation. There are two main reasons for this.
One is that even though there is no explicit loop closure, due to object and plane's long-range visibility properties, the algorithm may still associate with the old plane landmark to reduce the final drift. The second reason is that more feature points' depth can be initialized by object and planes especially when there is large camera rotations. Due to the strict outlier rejection and robust BA optimization, even if objects and planes don't improve the results, they won't seriously damage the system.

From the table, we also find that loop closure in ORB SLAM has some benefits in small office environments, but in large corridors with loops at the sequence end, it doesn't perform well compared to ours, because SLAM  already has large scale drift before the loop closure, the final global BA cannot fully recover the drift.
%sometimes there is no loop. or there is always large drift before loop closure. then no use any more. Overall, ORB with loop closure has some benefit in small environments such as office, but in large corridor dataset with only one loop at the end, it doesn't perform well compared to ours, because there is always large scale drift before the loop closure, global BA cannot fully correct them even with relocalization.

%Actually in the new experiment Table II, we find that our algorithm still performs better most of the time compared to ORB with loop closure. The main reason is that we evaluate the absolute trajectory error. So if there is already large drift, especially scale drift, before ORB SLAM loop closure, the final bundle adjustment cannot fully correct the trajectory. However if we evaluate trajectory after scale alignment, then ORB with loop closure will perform better.

\begin{table}[t]
\vspace{0.25em}
\caption{Absolute Camera Translation Error on Various Datasets ($cm$ for ICL, $m$ for others)}
\begin{center}
\begin{tabular}{c c c c}
\toprule
Method         & ORB \cite{mur2015orb}   & ORB-No LC \cite{mur2015orb}    & Ours  \\  \hline
ICL living 0   & 2.35 & 3.08    & \textbf{0.8} \\
ICL living 2   & 3.54 & 3.25    & \textbf{2.06} \\
ICL living 3   & \textbf{4.68} &5.36    & 5.38 \\
ICL office 0   & \textbf{5.67} & 6.23    & 5.93 \\
ICL office 2   & 3.82 & 5.00    & \textbf{2.63} \\  \hline

Tamu corridor  & 1.74 & 3.87    & \textbf{0.97} \\  \hline

Our room 1     & 0.14 & 0.15    & \textbf{0.05} \\
Our corridor 1 & 1.49 & 2.25    & \textbf{0.30} \\
Our corridor 2 & 1.05 & 2.93    & \textbf{0.24} \\
Our corridor 3 & 0.87 & 1.84    & \textbf{0.49} \\
\bottomrule
\end{tabular}
\end{center}
\label{table:ate pose error}
\end{table}

For TUM mono data in Table \ref{table:align pose error}, there is no ground truth camera height available thus we evaluate the monocular scale alignment error proposed in \cite{engel2017direct}. Results of DSO and ORB are taken from the supplementary material of DSO. Our semantic SLAM can work robustly in these challenging datasets even though there is large camera rotation and sometimes the camera may be upside down. In the cluttered dataset such as Room 37, there are only a few planes in a few observed frames thus our algorithm almost reduces to point SLAM and achieves. In Corridor 38, our algorithm and ORB SLAM are much worse compared to DSO because there are many areas of only one white wall with few feature points which are difficult for feature and plane based SLAM.

\begin{table}[t]
\caption{Pose Alignment Error on TUM-mono Dataset}
\begin{center}
\begin{tabular}{c c c c}
\toprule
Method        & ORB-No LC\cite{mur2015orb}   & DSO \cite{engel2017direct}  & Ours  \\  \hline
Corridor 36  & 1.81     & 4.01    & \textbf{0.94} \\
Room 37  & 0.60     & 0.55    & \textbf{0.35} \\
Corridor 38  & 23.9     & \textbf{0.55}    & 7.65 \\
\bottomrule
\end{tabular}
\end{center}
\label{table:align pose error}
\end{table}

\subsubsection{Time Analysis}
\label{sec: time analysis}

We also provide the run-time analysis on Intel i7-4790 CPU at 4.0 GHz and Nvidia 980 Ti GPU. GPU is used for 2D object detector and semantic segmentation. All SLAM parts are implemented in C++ on CPU. As shown in Table \ref{table:time analysis 2}, there are several single image pre-processing steps. The CNN algorithms we used cannot run in real time but they actually depend on the model complexity which can be replaced by recent lightweight CNNs.

The SLAM experiment runs on ICL-NUIM living room dataset. On average, there are 5 object landmarks in each local BA optimization. The tracking thread includes feature detection, associations, and camera pose tracking for each frame which can run in real time from the table. The BA map optimization occurs when a new keyframe is created, therefore it does not need to run in real-time. Compared to point only BA, adding objects into the system only increases the optimization time by $7\%$. Plane landmarks further double the optimization time because the point-plane constraints are applied to many points, bringing in more measurement costs to optimize. Another reason relates to the implementation of g2o. Since there are different types of edges with different dimensions such as camera-point, point-plane, we cannot pre-allocate the solver matrix dimensions.

%This is also reasonable as there are only a few objects in the local map optimization. 

%Since the number of objects and planes is far less compared to point features, the overall BA optimization is still efficient enough to run in real time.

\begin{table}[t]
\caption[Average runtime of different system components]{Average runtime of different SLAM components}
\begin{center}
\begin{tabular}{c |  c | c c}
\hline
\multirow{2}{*}{Dataset}  & \multirow{2}{*}{Tasks} 	&Runtime \\ 
						  &							& (mSec) \\ \hline
\multirow{3}{*}{\shortstack{Single image\\ Preprocessing}}  & Yolo 2D object detection    &17.5 \\
& SegNet semantic segmentation  &71.5 \\
		  	  & Edge detection	  &12.1 \\  \hline						  
\multirow{4}{*}{\shortstack{Indoor\\ ICL room}}   		&Tracking thread   		&15.0   \\ 
 	  	       &Point only BA   			&49.5   \\  
  			  &Point + object BA   	    &55.3   \\
  	     &Point + object + plane BA   &105.6   \\  \hline

\end{tabular}
\end{center}
\label{table:time analysis 2}
\end{table}

\section{Conclusion}
\label{sec:conclusion}
% propose what... results... future

In this work, we propose the first monocular SLAM and dense mapping algorithm combining points with high-level object and plane landmarks through unified BA optimization. We show that semantic scene understanding and traditional SLAM optimization can improve each other.

For the single image, we propose a fast 3D object and layout joint understanding for general indoor environments. Cuboid and plane proposals are generated from 2D object and edge detection. Then an efficient sparse high order CRF inference is proposed to select the best proposals. In the SLAM part, several new measurement functions are designed for planes and objects. Compared to points, objects and planes can provide long-range geometric and semantic constraints such as intersection and supporting relationships, to improve the pose estimation. Strict outlier rejection, robust data association and optimization are proposed to improve the robustness.

We evaluate the SLAM algorithm in various public indoor datasets including rooms and corridors. Our approach can improve the camera pose estimation and dense mapping in most environments compared to the state-of-the-art. 
%It reduces to a standard point SLAM if there are few objects and plane observed. 

In the future, more general planes in addition to wall planes need to be considered to produce a denser and more complete map. Dynamic objects and object surface mapping can also be addressed to improve the robustness and mapping quality.

\section*{APPENDIX}
\label{sec:appendix}
We here explain the CRF inference of Section \ref{sec:CRF inference} in more detail shown in Algorithm \ref{alg:crf message passing}. As mentioned before, there are $N+1$ special state for a clique with size $N$. For each state $\mathbf{y}_k$, we define $\mathbf{s}_k=\sum_{j\in \mathbf{y}_k}  m_{j\rightarrow c}^{t-1}(y_k^j)$ as the sum of messages to the clique. The key observation is that $\mathbf{s}_k$ can be computed iteratively and efficiently.

\begin{algorithm}[h]
\caption{CRF Factor-to-Variable Message Passing}
\SetAlgoLined
\KwIn{Variable-to-factor message $m_{i\rightarrow c}^{t-1}(\cdot), i=1,...,N$}
\KwOut{All Factor-to-variable message: $m_{c\rightarrow i}^{t}(x_i), i=1,...,N$}

// compute all $\mathbf{s}_k$ recursively and record min \linebreak
$\mathbf{s}_1=\sum_{j=1,2,3...N} m_{j\rightarrow c}^{t-1}(\mathbf{y}_1^j)$  \linebreak
$\text{minS} = \mathbf{s}_1;\text{sndminS} = \infty$  \linebreak
\For{$k\leftarrow 2$ \KwTo $N+1$}{
\eIf{$k \leq N$}
{
$\mathbf{s}_k = \mathbf{s}_{k-1}-m_{k-1\rightarrow c}^{t-1}(1) + m_{k-1\rightarrow c}^{t-1}(0) - m_{k\rightarrow c}^{t-1}(0) + m_{k\rightarrow c}^{t-1}(1)$    
}
{
$\mathbf{s}_k = \mathbf{s}_{k-1}-m_{k-1\rightarrow c}^{t-1}(1) + m_{k-1\rightarrow c}^{t-1}(0)$
}
\eIf{$\mathbf{s}_k \leq  \text{minS}$}
{
	$\text{minS} = \mathbf{s}_k$
}
{
	\If{$\mathbf{s}_k \leq \text{sndminS}$}
	{
		$\text{sndminS} = \mathbf{s}_k$
	}
}
}
// compute final message \linebreak
\For{$k\leftarrow 1$ \KwTo $N$}{
    $m_{c\rightarrow k}^t (x_k=1) = \mathbf{s}_k- m_{k\rightarrow c}^{t-1}(x_k=1)$ \linebreak
	\eIf{$\mathbf{s}_k == \text{minS}_{\text{val}}$}
	{
		$m_{c\rightarrow k}^t (x_k=0) = \text{sndminS} - m_{k\rightarrow c}^{t-1}(x_k=0)$
	}
	{
		$m_{c\rightarrow k}^t (x_k=0) = \text{minS} - m_{k\rightarrow c}^{t-1}(x_k=0)$
	}
}

\label{alg:crf message passing}
\end{algorithm}

%\section*{ACKNOWLEDGMENT}

\balance

\bibliographystyle{unsrt}    % reference order according to appearance
\bibliography{ref}

%\addtolength{\textheight}{-12cm}   % This command serves to balance the column lengths
                                  % on the last page of the document manually. It shortens
                                  % the textheight of the last page by a suitable amount.
                                  % This command does not take effect until the next page
                                  % so it should come on the page before the last. Make
                                  % sure that you do not shorten the textheight too much.

\end{document}